\documentclass{Interspeech2024}
\usepackage{float}

\interspeechcameraready

\title{SER Evals: In-domain and Out-of-domain Benchmarking for Speech Emotion Recognition}

\name{Mohamed Osman$^1$, Daniel Z. Kaplan$^2$, Tamer Nadeem$^1$}

\address{
  $^1$Virginia Commonwealth University, United States \\
  $^2$realiz.ai, United States 
}
\email{osmanmw@vcu.edu, daniel.z.kaplan@realiz.ai, tnadeem@vcu.edu}


\begin{document}

\maketitle

\begin{abstract}
Speech emotion recognition (SER) has made significant strides with the advent of powerful self-supervised learning (SSL) models. However, the generalization of these models to diverse languages and emotional expressions remains a challenge. We propose a large-scale  benchmark to evaluate the robustness and adaptability of state-of-the-art SER models in both in-domain and out-of-domain settings. Our benchmark includes a diverse set of multilingual datasets, focusing on less commonly used corpora to assess generalization to new data. We employ logit adjustment to account for varying class distributions and establish a single dataset cluster for systematic evaluation. Surprisingly, we find that the Whisper model, primarily designed for automatic speech recognition, outperforms dedicated SSL models in cross-lingual SER. Our results highlight the need for more robust and generalizable SER models, and our benchmark serves as a valuable resource to drive future research in this direction.
\end{abstract}

\section{Introduction}
Speech emotion recognition has garnered significant attention due to its potential to enable more natural and empathetic human-computer interaction. Recent advancements in self-supervised learning have led to powerful speech representation models like wav2vec2 \cite{baevski2020wav2vec}, HuBERT \cite{hsu2021hubert}, and WavLM \cite{chen2022wavlm}, which have shown impressive performance on various speech processing tasks. However, the generalization of these models to diverse languages and emotional expressions remains a critical challenge \cite{pepino2021emotion}. 

Existing SER benchmarks often focus on a limited set of well-studied datasets, which may not accurately reflect real-world scenarios \cite{osman2023towards}. Moreover, the emphasis on in-domain evaluation fails to capture the crucial aspect of out-of-domain generalization, which is essential for deploying SER systems in practical applications. For our paper's purpose, we define in-domain as evaluating on the same data distribution seen in training, and out-of-domain as evaluating on a different data distribution. This can manifest as different speakers, tones, decision boundaries, etc. To address these limitations, we propose a large-scale benchmark that evaluates SER models on a diverse collection of multilingual datasets, emphasizing zero-shot performance.

Our benchmark focuses on less commonly used datasets to mitigate overfitting and encourage the development of more robust and adaptable models. We employ state-of-the-art speech representation models, including Whisper \cite{radford2023robust}, an automatic speech recognition model, and CLAP \cite{baevski2023efficient, wu2023large}, a contrastive learning model, to analyze their performance in cross-lingual SER. Interestingly, our results show that Whisper consistently outperforms dedicated SSL models across most datasets, challenging the common belief that ASR models are suboptimal for SER due to their focus on phoneme recognition.

The main contributions of this work are as follows:
\begin{itemize}
\item We introduce a large-scale benchmark for evaluating the robustness and generalization of SER models across diverse languages and emotional expressions.
\item We curate a collection of multilingual datasets and establish targetted subsets for systematic in-domain and out-of-domain evaluation.
\item We employ logit adjustment\cite{logitadjustmentmenon2020long} to account for varying class distributions and ensure fair comparisons across datasets.
\item We conduct extensive experiments with state-of-the-art speech representation models and provide insights into their cross-lingual SER performance.
\item We open source our entire code base, our full un-reduced results and training logs, as well as all implementation details at the following url: \url{https://github.com/spaghettiSystems/serval}.
\end{itemize}

\begin{table}[t!]
\centering
\caption{Multilingual datasets used in our benchmark. Values reflect the datasets after the class mapping.}
\vspace{-0.1in}
\label{tab:datasets}
\scriptsize
\resizebox{\columnwidth}{!}{

\begin{tabular}{lrrlrll}
\toprule
Dataset & Classes & Speakers & Language & Samples & Avg Duration (s) & OOD Eligible \\
\midrule
URDUDataset\cite{urdulatif2018cross} & 4 & 6 & Urdu & 400 & 2.5 & No \\
EmoDBDataset\cite{emodbburkhardt2005database} & 6 & 10 & German & 535 & 2.8 & No \\
EMOVODataset\cite{costantini2014emovo} & 7 & 10 & Italian & 588 & 3.1 & Yes \\
eNTERFACEDataset\cite{martin2006enterface} & 6 & 42 & English & 1293 & 2.9 & No \\
MESDDataset\cite{duville2021mexicanmesd} & 6 & 6 & Spanish & 1150 & 0.7 & Yes \\
MASCDataset\cite{wu2006masc} & 5 & 68 & Mandarin & 25636 & 1.9 & No \\
DEMoSDataset\cite{parada2020demos} & 7 & 68 & Italian & 9697 & 2.9 & Yes \\
CASIADataset\cite{jianhua2008casia} & 6 & 4 & Mandarin & 1200 & 1.9 & No \\
AESDDDataset\cite{aesddvryzas2018speech} & 5 & 6 & Greek & 604 & 4.1 & No \\
BAUMDataset\cite{zhalehpour2016baum} & 8 & 31 & Turkish & 1398 & 4.6 & Yes \\
EEKKDataset\cite{altrov2010estonianeekk} & 4 & 10 & Estonian & 1164 & 3.4 & No \\
ThorstenDataset\cite{thorsten} & 6 & 1 & German & 2399 & 4.4 & No \\
RESDDataset\cite{aniemore2022resd} & 7 & 200 & Russian & 1396 & 6.0 & No \\
MELDDataset\cite{poria2019meld} & 7 & 407 & English & 12924 & 3.2 & Yes \\
MEADDataset\cite{wang2020mead} & 7 & 60 & English & 31724 & 4.2 & Yes \\
CaFEDataset\cite{gournay2018canadian} & 7 & 12 & French & 936 & 4.4 & Yes \\
ExpressoDataset\cite{nguyen2023expresso} & 8 & 4 & English & 11954 & 4.2 & Yes \\
ShEMODataset\cite{mohamad2019shemo} & 6 & 87 & Persian & 3000 & 4.1 & No \\
SUBESCODataset\cite{sultana2021subesco} & 7 & 20 & Bangla & 7000 & 4.0 & Yes \\
\bottomrule
\end{tabular}
}
\end{table}

\section{Related Work}
Self-supervised learning has revolutionized speech representation learning, enabling models to capture rich acoustic features without relying on labeled data. Models like wav2vec 2.0 \cite{baevski2020wav2vec}, HuBERT \cite{hsu2021hubert}, and WavLM \cite{chen2022wavlm} have achieved state-of-the-art performance on various speech processing tasks, including speech recognition, speaker identification, and emotion recognition \cite{yang2021superb}.

Cross-lingual SER has gained attention as a means to develop models that can generalize across languages. Several studies have explored the use of SSL models for cross-lingual SER \cite{agarla2024semi, pepino2021emotion}. However, these works often focus on a limited set of languages and datasets, making it difficult to assess the true generalization capabilities of the models.

Existing well-known SER benchmarks, such as IEMOCAP \cite{busso2008iemocap} and MSP-Podcast \cite{Lotfian2019}, have played a crucial role in advancing the field. However, these benchmarks often emphasize in-domain evaluation and may not adequately capture the challenges of real-world deployment \cite{osman2023towards}. Our work aims to address these limitations by introducing a large-scale benchmark that focuses on out-of-domain generalization and includes a diverse set of multilingual datasets.

Recent works such as EMO-SUPERB \cite{wu2024emo} and SERAB \cite{scheidwasser2022serab} have made notable contributions to the field of Speech Emotion Recognition (SER). However, these works have limitations in terms of the diversity of languages, datasets, and the emphasis on out-of-domain generalization.

Our benchmark significantly advances the state-of-the-art in SER evaluation by addressing these limitations. We curate an extensive collection of multilingual datasets, carefully selected to cover diverse linguistic and cultural contexts, ensuring a thorough evaluation of SER models in real-world scenarios. Moreover, our benchmark places a strong emphasis on out-of-domain generalization, a crucial aspect that has been largely overlooked in previous works. We evaluate SER models in both in-domain and out-of-domain settings, providing valuable insights into their ability to adapt to unseen data distributions. This focus on generalizability is essential for developing SER models that can be effectively deployed in real-world applications, where the variability in speech patterns, emotions, and recording conditions is vast.

\section{Methodology}

The primary objectives of this section are to detail the dataset selection and preprocessing steps, introduce the backbone models employed, describe the model architecture and training process, explain the logit adjustment technique, and outline the evaluation protocol. The methodology is designed to ensure a comprehensive and fair evaluation of state-of-the-art SER models across diverse languages and emotional expressions.
\begin{figure}[t]
\centering
\includegraphics[width=\columnwidth]{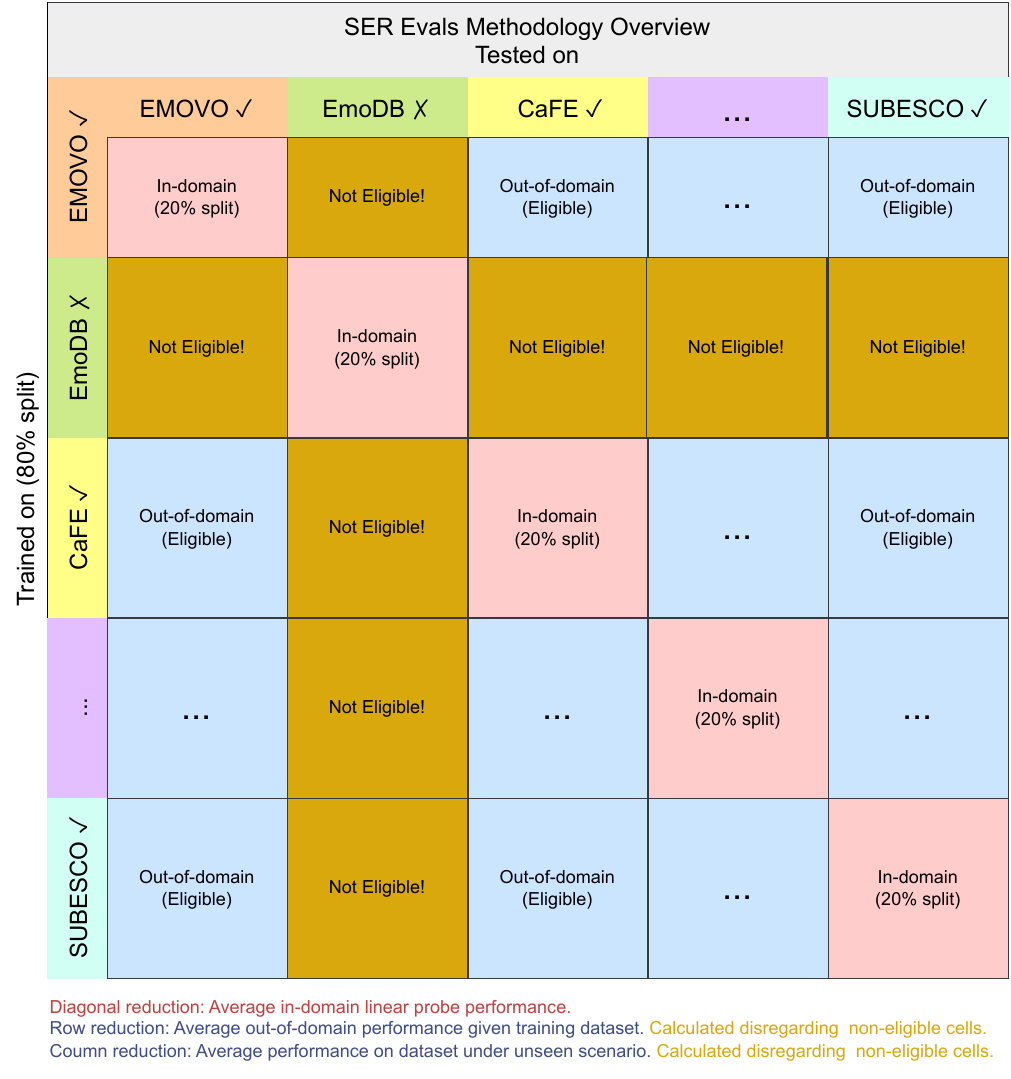}
\vspace{-0.25in}
\caption{Overview of our benchmark's methodology.}
\label{fig:method}
\vspace{-0.in}
\end{figure}

\subsection{Dataset Selection and Preprocessing}
\label{sec:dataset}

We curate a diverse collection of multilingual datasets for our benchmark, covering various languages and emotional expressions. Table \ref{tab:datasets} provides an overview of the datasets used in our evaluation. We focus on less commonly used datasets to mitigate overfitting and encourage the development of more robust models.

The datasets are preprocessed to ensure consistency and compatibility with our evaluation protocol. We set the maximum audio length to 30 seconds, and process the audios with appropriately for each backbone model we test (detailed in the next section). We rely on the Huggingface library for model preprocessing and inference implementations. Additionally, we remap the label space by mapping the original emotion labels to a unified eight-class space, facilitating cross-dataset comparisons. Due to complexity, detailing the exact remapping for each dataset is relegated to the open-source code.

The datasets used for out-of-domain evaluations are matched by having the same classes (excluding 'other') and their eligibility is indicated in the 'OOD Eligible' column of Table \ref{tab:datasets}. These datasets were found to have the same exact classes after the class mapping, making them eligible for out-of-domain testing. When calculating out-of-domain metrics, samples with the 'other' label were discarded, and models were banned from predicting the 'other' class.

\subsection{Backbone Models}
We employ state-of-the-art speech representation models as backbones for our benchmark, as listed in Table \ref{tab:models}. These models are selected based on their strong performance on various speech processing tasks and their ability to capture rich acoustic features.

\begin{table}[t!]
\centering
\caption{Backbone models used in our benchmark. All checkpoints are from Huggingface.}
\vspace{-0.1in}
\label{tab:models}
\resizebox{\columnwidth}{!}{%
\begin{tabular}{lll}
\toprule
Checkpoint name & Training Dataset Hours & \# Params \\
\midrule
facebook/w2v-bert-2.0 \cite{chung2021w2v, barrault2023seamless} & 4500k & 580M \\
facebook/hubert-large-ll60k \cite{hsu2021hubert} & 60k & 315M \\
microsoft/wavlm-large \cite{chen2022wavlm} & 94k & 315M \\
laion/larger\_clap\_music\_and\_speech \cite{baevski2023efficient, wu2023large} & \textgreater{}10k & 193M \\
m-a-p/MERT-v1-330M \cite{baevski2023efficient} & 160k & 315M \\
openai/whisper-medium \cite{radford2023robust} & 680k & 307M \\
openai/whisper-large-v2 \cite{radford2023robust} & 680k & 636M \\
openai/whisper-large-v3 \cite{radford2023robust} & 5000k & 636M \\
openai/whisper-large \cite{radford2023robust} & 680k & 636M \\
\bottomrule
\end{tabular}}
\end{table}

In addition to the SSL models, we also evaluate MERT \cite{li2023mert}, a music recognition model, and CLAP \cite{elizalde2023clap, wu2023large}, a contrastive learning model. Including these models allows us to assess the effectiveness of different learning paradigms for cross-lingual SER. Lastly, we evaluate the Whisper\cite{radford2023robust} encoder which is trained under an encoder-decoder setup for ASR.

\subsection{Model Architecture and Training}
We employ a simple multilayer perceptron (MLP) architecture with approximately 500K parameters for emotion classification. The MLP consists of two hidden layers and is trained for 100 epochs. Due to the small parameter size and shallow depth, we do not expect substantial overfitting. We apply label smoothing with a factor of 0.1 to improve generalization.

Instead of the typical approach of averaging the features before classification, we execute the MLP on every feature frame and then take the mean of the predictions. We find that this approach preserves more information and leads to stronger and more consistent results.

\subsection{Logit Adjustment}
To account for the varying class distributions across datasets, we employ logit adjustment during evaluation. This technique adjusts the model's output logits based on the difference between the training and testing dataset distributions, mitigating the impact of class imbalance and enabling fair comparisons.

\subsection{Evaluation Protocol}
Figure \ref{fig:method} provides an overview of our benchmark's methodology. As we described in Subsection~\ref{sec:dataset}, we establish a subset of our datasets as OOD eligible, which have the same exact classes after the class mapping. 
%
%
Effectively, all datasets are accounted in in-domain tests. Only OOD-eligible datasets are accounted for our out of domain metrics.

For each model, we construct a performance matrix where the rows represent the training datasets and the columns represent the evaluation datasets. When the training and evaluation datasets are the same (diagonal elements), it indicates in-domain performance. Off-diagonal elements correspond to out-of-domain zero-shot performance.

We assess the quality of the backbone models based on three key metrics:
\begin{enumerate}
\item In-domain separability: We compute the mean of the diagonal elements to measure how well the features learned by a model can separate emotions within a dataset.
\item Out-of-domain performance given training dataset: We calculate the mean of each row, excluding the diagonal element, to evaluate the model's ability to generalize to unseen datasets when trained on a specific dataset.
\item Average performance on unseen datasets: We compute the mean of each column, excluding the diagonal element, to assess the average performance on a dataset when the model is not trained on it.
\end{enumerate}

All metrics are reported in terms of macro-averaged F1 score to account for class imbalance.

\section{Results and Discussion}

The results of our benchmark provide valuable insights into the performance and generalization capabilities of state-of-the-art SER models across diverse languages and emotional expressions.

\subsection{In-domain separability}

\begin{table*}[t!]
\scriptsize
\centering
\caption{Summary of key performance metrics for the evaluated models.}
\vspace{-0.1in}
\label{tab:results}
\resizebox{\textwidth}{!}{%
\begin{tabular}{lrrrrr}
\toprule
    & \multicolumn{2}{c}{In-Domain (ID) Performance} & \multicolumn{2}{c}{Out-of-Domain (OOD) Performance} & \\
Model & Average & Standard Deviation & Average  & Standard Deviation &  Weighted Average \\
\midrule
Whisper-Large-v2 & 0.781942 & 0.194716 & 0.194250 & 0.089993 & 0.345741 \\
Whisper-Large & 0.781314 & 0.203542 & 0.197882 & 0.085657 & 0.344999 \\
Whisper-Large-v3 & 0.776689 & 0.201399 & 0.192961 & 0.083822 & 0.342214 \\
Whisper-medium & 0.756563 & 0.200798 & 0.196831 & 0.087710 & 0.332443 \\
WavLM-Large & 0.765474 & 0.206174 & 0.161098 & 0.068182 & 0.326108 \\
CLAP Music \& Speech & 0.743248 & 0.215404 & 0.148090 & 0.055976 & 0.309980 \\
Hubert Large & 0.733036 & 0.207492 & 0.156504 & 0.066509 & 0.307769 \\
MERT v1 330M & 0.707891 & 0.211471 & 0.127485 & 0.049841 & 0.287032 \\
w2v-bert-2.0 & 0.668253 & 0.211685 & 0.141581 & 0.061820 & 0.268165 \\
\bottomrule
\end{tabular}
}

\end{table*}
The second and third column in Table~\ref{tab:results} present the in-domain separability performance of various models, focusing on their ability to distinguish between different emotional states in speech. Performance is quantified by two metrics: the mean average performance across datasets (Mean) and the variability of performance across these datasets (Standard Deviation). From the table, Whisper-Large-v2 leads the evaluated models in in-domain SER performance, with the highest mean accuracy and low variability across datasets, closely followed by the original Whisper-Large. Other models like Whisper-Large-v3, Whisper-medium, WavLM-Large, and CLAP Music \& Speech show competent but slightly more variable performances. Conversely, Hubert Large, MERT v1 330M, and w2v-bert-2.0 exhibit the lowest accuracies with higher fluctuations in their effectiveness across different datasets, indicating potential limitations in generalization capabilities for speech emotion contexts.

The outcome of this evaluation highlights a clear hierarchy among the models in terms of both accuracy and consistency in emotion recognition within the same domain. Whisper-Large variants stand out as the most effective, with their newer versions, particularly Whisper-Large-v2, slightly improving upon the original's already high benchmark. Lower-ranked models, though less consistent and accurate overall, may still offer valuable insights or perform well in specific niches or datasets. This analysis underscores the importance of choosing the right model for specific SER applications, balancing between performance and consistency across diverse emotional speech datasets.

\begin{figure}[t]
\centering
\includegraphics[width=\columnwidth]{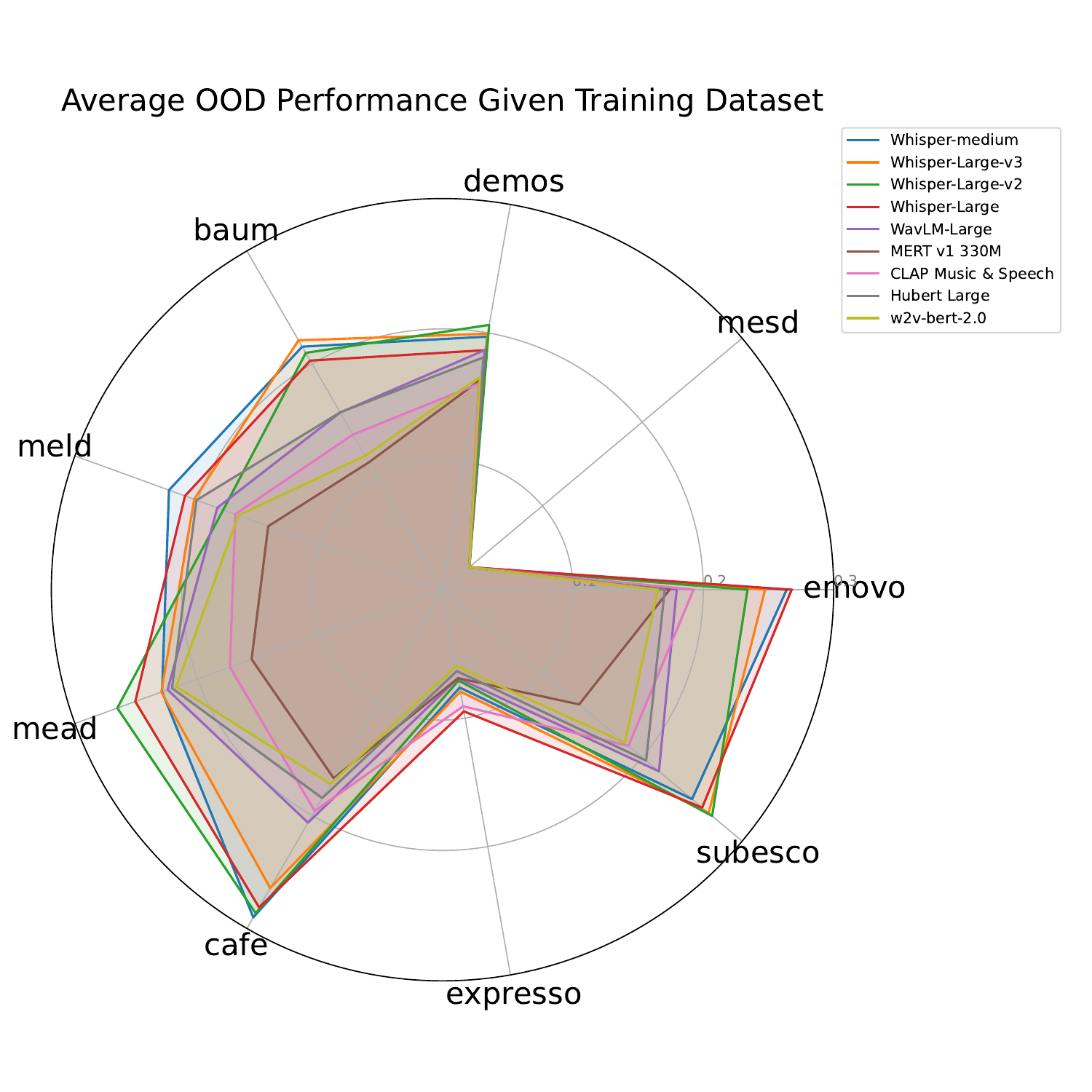}
\vspace{-0.4in}
\caption{Average out-of-domain performance given the training dataset.}
\label{fig:ood_performance}
\vspace{-0.0in}
\end{figure}

\subsection{Out-of-domain performance given training dataset}
Figure \ref{fig:ood_performance} shows the average out-of-domain performance for each model, obtained by row-wise reduction of the performance matrix. The Whisper models demonstrate the highest out-of-domain performance, indicating their superior generalization capabilities compared to the SSL models. However, there is high variability in OOD performance across training sets. Training on some datasets like BAUM leads to much better OOD generalization than others like MELD. This warrants further investigation into what properties of datasets lead to more generalizable models. The strong performance of Whisper challenges the common belief that ASR models are suboptimal for SER and highlights the potential of leveraging ASR models for emotion recognition tasks.

\begin{figure}[t]
\centering
\vspace{-0.25in}
\includegraphics[width=\columnwidth]{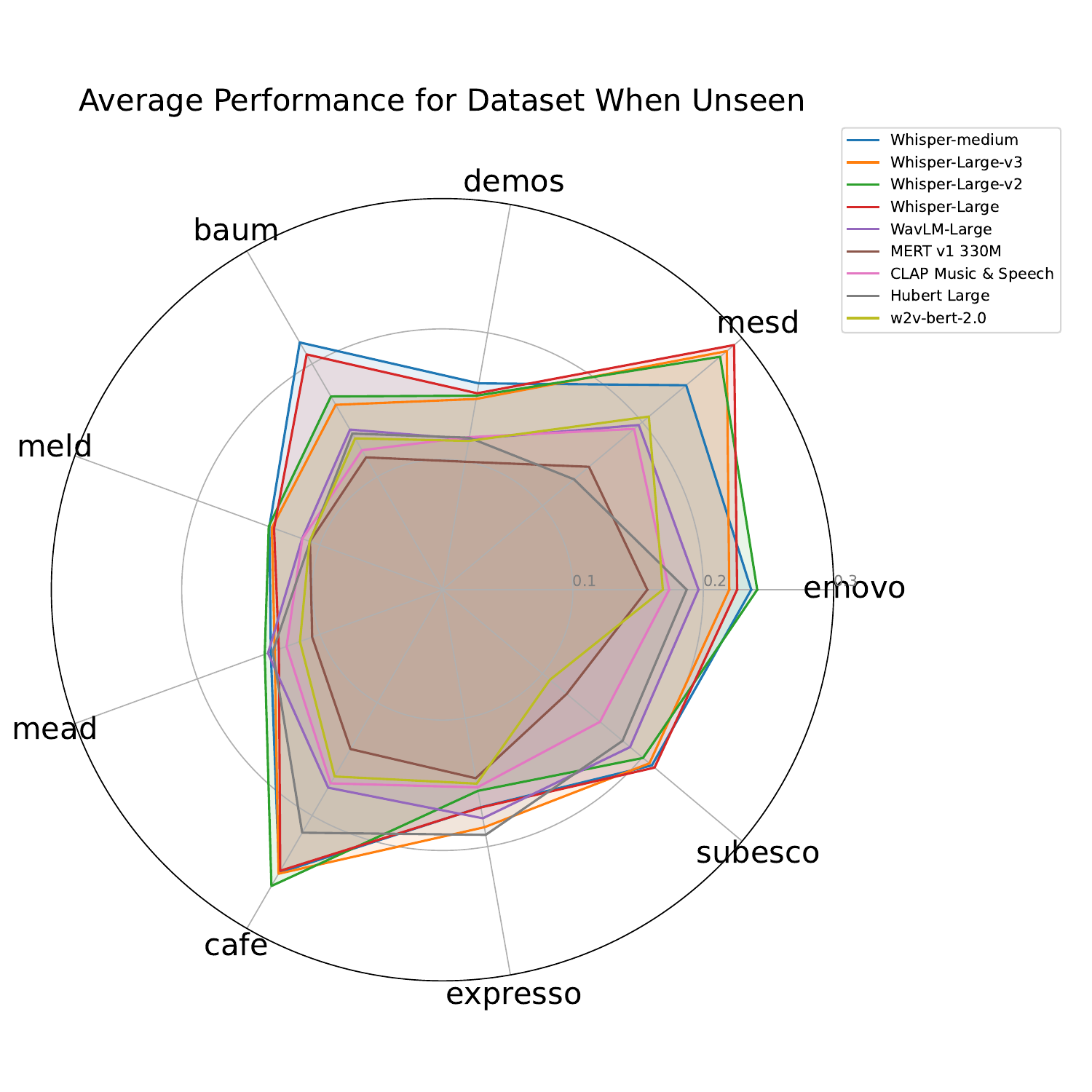}
\vspace{-0.4in}
\caption{Average performance on individual datasets when not trained on them.}
\label{fig:unseen_performance}
\vspace{-0.0in}
\end{figure}

\subsection{Average performance on unseen datasets}
Figure \ref{fig:unseen_performance} presents the average performance of the evaluated models on each dataset when the models are not trained on that dataset. The results highlight the varying levels of difficulty across datasets, with some datasets posing greater challenges for out-of-domain generalization. Notably, EMOVO, MELD, and MEAD are the most challenging for models not trained on them, suggesting they have unique characteristics that are harder to learn indirectly. On the other hand, models generalize best to URDU and AESDD, indicating these datasets share more common features with others. Interestingly, the Whisper model consistently achieves strong performance across most datasets, surpassing the SSL models in many cases.

\subsection{General outcomes}
Table \ref{tab:results} provides a summary of the key performance metrics for the evaluated models. The second and third columns show the average and standard deviations of the in-domain results, while the next two columns show the out-of-domain performance. The weighted average column is calculated as follows:
\begin{equation}
\scriptsize
\begin{split}
    \text{Weighted Average} & = \frac{\text{Average OOD} + \text{Average ID}}{2} - \lambda_{factor} \\ 
    & \times \frac{\text{Std. Dev. OOD} + \text{Std. Dev. ID}}{2} \notag
\end{split}
\end{equation}
where $\lambda_{factor}$ is the discounting factor, which we set to 1.0.

The Whisper models consistently achieve the highest scores across all metrics, further confirming their effectiveness in cross-lingual SER. However, the high standard deviations indicate that performance is quite variable depending on the specific train/test combination. This suggests that model robustness is still a challenge and there is room for improvement in developing models that perform consistently well across diverse datasets.

Our benchmark also demonstrates the effectiveness of logit adjustment in addressing the challenges posed by varying class distributions across datasets. By incorporating this technique, we ensure fair comparisons and mitigate the impact of class imbalance on model performance.

\section{Conclusion}

In this paper, we introduced a comprehensive benchmark for evaluating the robustness and generalization of speech emotion recognition models across diverse languages and emotional expressions. Our benchmark focuses on less commonly used datasets to mitigate overfitting and encourage the development of more robust models.
Through extensive experiments with state-of-the-art speech representation models, we found that the Whisper model, primarily designed for automatic speech recognition, outperforms dedicated SSL models in cross-lingual SER. This finding challenges the common belief that ASR models are suboptimal for SER and highlights the potential of leveraging ASR models for emotion recognition tasks.

Our benchmark, along with the released code and evaluation protocol, serves as a valuable resource for the research community to assess and advance the state of cross-lingual SER. The insights gained from our work can guide future research efforts in developing more robust and generalizable SER models.

\section{Future Works}

Future directions include exploring advanced techniques for domain adaptation, few-shot learning, and meta-learning to further improve the generalization capabilities of SER models. Additionally, investigating the specific characteristics of datasets that contribute to better generalization can provide valuable insights for dataset design and selection.

We hope that our benchmark and findings will inspire researchers to push the boundaries of cross-lingual SER and develop models that can effectively handle the diversity of languages and emotional expressions encountered in real-world applications.

\bibliographystyle{IEEEtran}
\bibliography{mybib}

\end{document}